\documentclass[nohyperref]{article}

\usepackage{microtype}
\usepackage{graphicx}
\usepackage{subfigure}
\usepackage{booktabs} 

\usepackage{hyperref}

\usepackage[accepted]{icml2022}

\usepackage{amsmath}
\usepackage{amssymb}
\usepackage{mathtools}
\usepackage{amsthm}

\usepackage[capitalize,noabbrev]{cleveref}

\theoremstyle{plain}

\theoremstyle{definition}

\theoremstyle{remark}

\usepackage[textsize=tiny]{todonotes}

\icmltitlerunning{Similarity of Pre-trained and Fine-tuned Representations}

\begin{document}

\twocolumn[
\icmltitle{Similarity of Pre-trained and Fine-tuned Representations}




\begin{icmlauthorlist}
\icmlauthor{Thomas Goerttler}{yyy}
\icmlauthor{Klaus Obermayer}{yyy,comp}
\end{icmlauthorlist}

\icmlaffiliation{yyy}{Chair of Neural Information Processing, Technische Universitat Berlin, Germany}
\icmlaffiliation{comp}{Bernstein Center for Computational Neuroscience Berlin, Germany}

\icmlcorrespondingauthor{Thomas Goerttler}{thomas.goerttler@tu-berlin.de}

\icmlkeywords{Transfer Learning, cross-domain adaption}

\vskip 0.3in
]



\printAffiliationsAndNotice{\icmlEqualContribution} 

\begin{abstract}
In transfer learning, only the last part of the networks - the so-called head - is often fine-tuned. Representation similarity analysis shows that the most significant change still occurs in the head even if all weights are updatable. However, recent results from few-shot learning have shown that representation change in the early layers, which are mostly convolutional, is beneficial, especially in the case of cross-domain adaption. In our paper, we find out whether that also holds true for transfer learning. In addition, we analyze the change of representation in transfer learning, both during pre-training and fine-tuning, and find out that pre-trained structure is \textit{unlearned} if not usable.
\end{abstract}

\section{Introduction}

Pre-trained weights are often reused when deep learning models are trained. There are several reasons to use weights from a model having been pre-trained on a larger dataset: the training is faster, less time is needed, the computational costs are decreased, and the performance improves, especially when only having a small dataset. For computer vision, using on ImageNet pre-trained weights have become the standard practice, as the weights encode information coming from millions of images in diverse domains.

However, it is still not fully revealed why transfer learning works so well. It is assumed that the first part of the network is only dealing with low-level features (e.g., edges and blobs) and, later on, high-level features (e.g., objects) \citep{DBLP:journals/pieee/ZhuangQDXZZXH21}. Therefore, only fine-tuning the head (which are the fully connected layers at the end) is often enough to adapt to the novel but still similar tasks. This is also an implicit regularization as reducing the parameter also prevents overfitting. \citep{DBLP:conf/iclr/RaghuRBV20} discovered that in a few-shot learning problem, a meta-learner trained to adapt fast to a novel task hardly changes the representation of early layers during fine-tuning if the task comes from the same distribution as the training tasks. However, \citet{DBLP:conf/iclr/OhYKY21} found out that, especially in the case of cross-domain adaption, where the fine-tuning task does not come from the same distribution as in training, also an adaptation of earlier layers is very beneficial. 
\citet{DBLP:conf/nips/NeyshaburSZ20} investigated what is transferred in transfer learning by shuffling the blocks of inputs. They confirmed that lower layers are responsible for more general features and that a network with pre-trained weights stays in the same basin of solution during fine-tuning.

This paper analyses representation obtained by models having initialized, pre-trained, and fine-tuned weights. We compare their corresponding representation and analyze their similarity when applying them to structured, unstructured, domain, and cross-domain tasks. We find out that structure in the data is encoded in the early convolutional layers. If the transfer task is unstructured, it \textit{unlearns} the information. If the data comes from the same domain or a cross-domain, it exploits it. 

\section{Similarity Analysis of Representation}

To analyze what happens during pre-training and fine-tuning, we compare the representation similarity of different layers of the model. 
Representation similarity techniques are widely used in computational neuroscience and machine learning.
In neuroscience, most often the RSA (representation similarity analysis) is used, e.g., to compare a computational or behavioral model with the brain response \citep{kriegeskorte2008representational}. In deep learning,  most often centered kernel alignment (CKA) \citep{DBLP:conf/icml/Kornblith0LH19} is used to compare the representation \citep{DBLP:conf/iclr/RaghuRBV20, DBLP:conf/iclr/OhYKY21, DBLP:conf/nips/KornblithCLN21, DBLP:conf/nips/NeyshaburSZ20}.
RSA and CKA are similar and are instances of a more general approach discussed by \citet{DBLP:conf/eccv/DwivediHCR20}.
In a first step, representations associated with each pair of inputs - which can be normalized - are compared (e.g., by Euclidean distances or calculating the scalar product). In a second step, the resulting Gram matrices of a layer can be compared to other gram matrices with another similarity measure. 

In our experiments, we use the 4-block-conv architecture which has been proposed by \citet{DBLP:conf/nips/VinyalsBLKW16} and also used in \citet{DBLP:conf/iclr/OhYKY21}. The architecture consists of 4 modules with 3x3 convolutions and 64 filters. These are followed by batch normalization, a ReLU activation function, and a pooling layer (2x2). The output of the fourth block is flattened and fully connected with the output layer. We use the stochastic gradient optimizer with a learning rate of $0.001$ and a momentum of $0.9$.

We apply the network to the two standard datasets, CIFAR-10 and SVHN. Every run is repeated five times (each with a different seed), and the standard error of the results is indicated.

\begin{figure}[ht]
\begin{center}
\centerline{\includegraphics[width=\columnwidth]{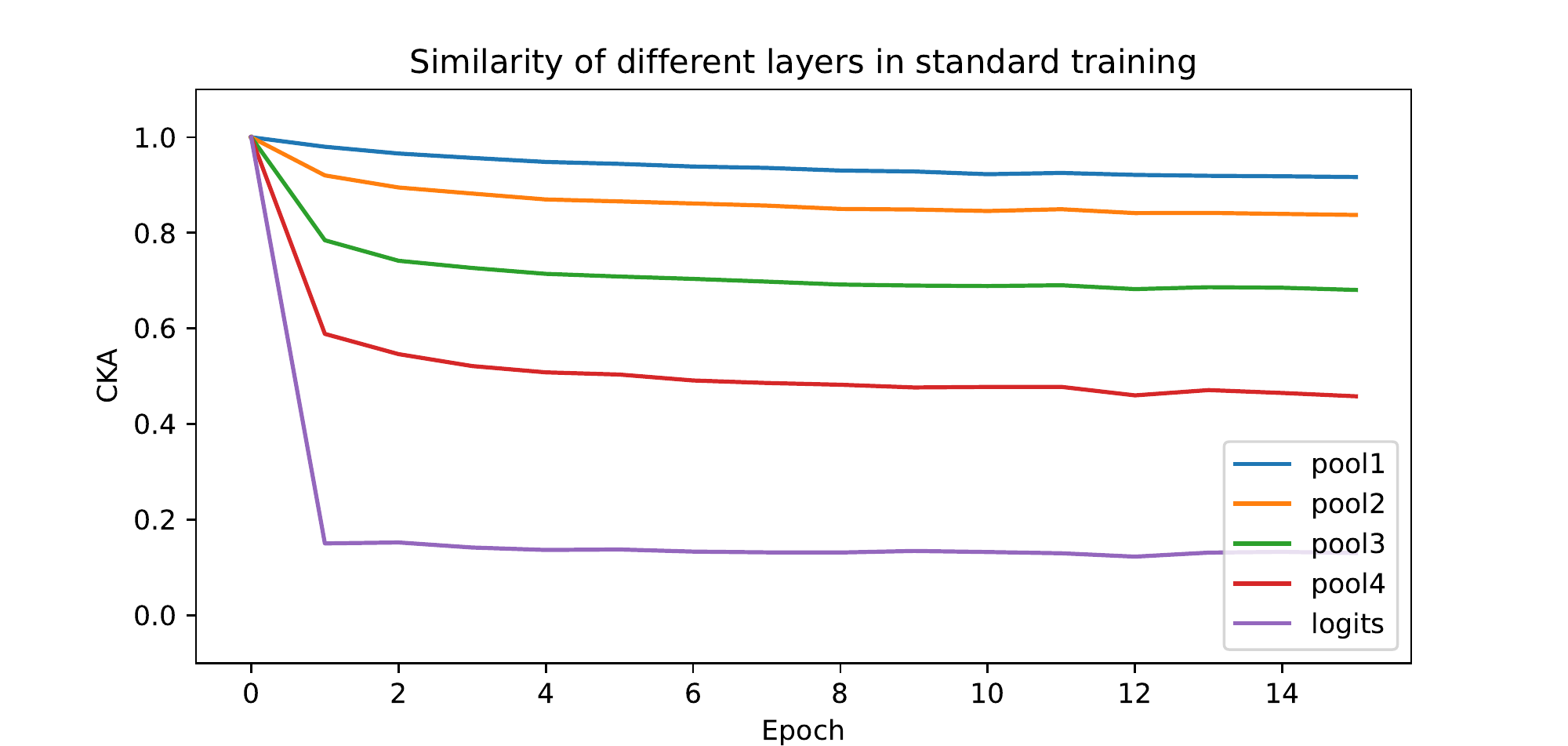}}
\caption{Similarity of the representations of the the 4-block-conv architecture applied on CIFAR-10 during training.}
    \label{fig:exp134}
\end{center}
\end{figure}

\section{Pre-trained Networks}\label{sec:pre-trained}

To understand how transfer learning learns, we first need to analyze what is learned in the standard deep learning, which serves in transfer learning as pre-training.
This gives a good overview of how networks generally learn and also the option to be directly compared to fine-tuning.

In our first experiment, we look at the representation change of the layers to the random initialization and control how it evolves during training (see Figure~\ref{fig:exp134}).
Similar to \citet{DBLP:journals/corr/abs-2105-05757}, we confirm that the representations of early layers change less than the later ones. This implies that less adaptation in the early layer is common for neural networks. Therefore, one has to be careful to interpret smaller adaptations in early layers directly but instead has to compare them always with another setup.

\subsection{Label Generation}

Since the scalar value of the similarity score of CKA is difficult to classify, we propose an experiment on random labels. For that, we also introduce partially random labels. If we have a labeled dataset with $D$ classes which are named with numbers from $0$ to $D-1$, we define a partially random label $y_d$ with a degree of randomness $d \in \{0, 1, ..., D-1\}$ as:
\begin{equation}
y_d = (y + Y) \mod D
\end{equation}
where $Y$ is a random variable with a probability function of
\begin{equation}
Pr(y=Y)=\begin{dcases}
        \frac{1}{d+1}& \text{if } y \in \mathbb{N}_0 \text{ and } y \leq d \\
        0& \text{otherwise} \\
    \end{dcases}
\end{equation}

Every original label is uniformly distributed over $d+1$ labels, such that the closer $d$ is to $D$, the more random the labels are, and less structure remains in the problem. A choice of $d=0$ is equivalent to the original dataset, whereas $d=D-1$ means that the dataset is completely random.

\begin{figure}[ht]
\centering     
\subfigure[]{\label{fig:loss}\includegraphics[width=\columnwidth]{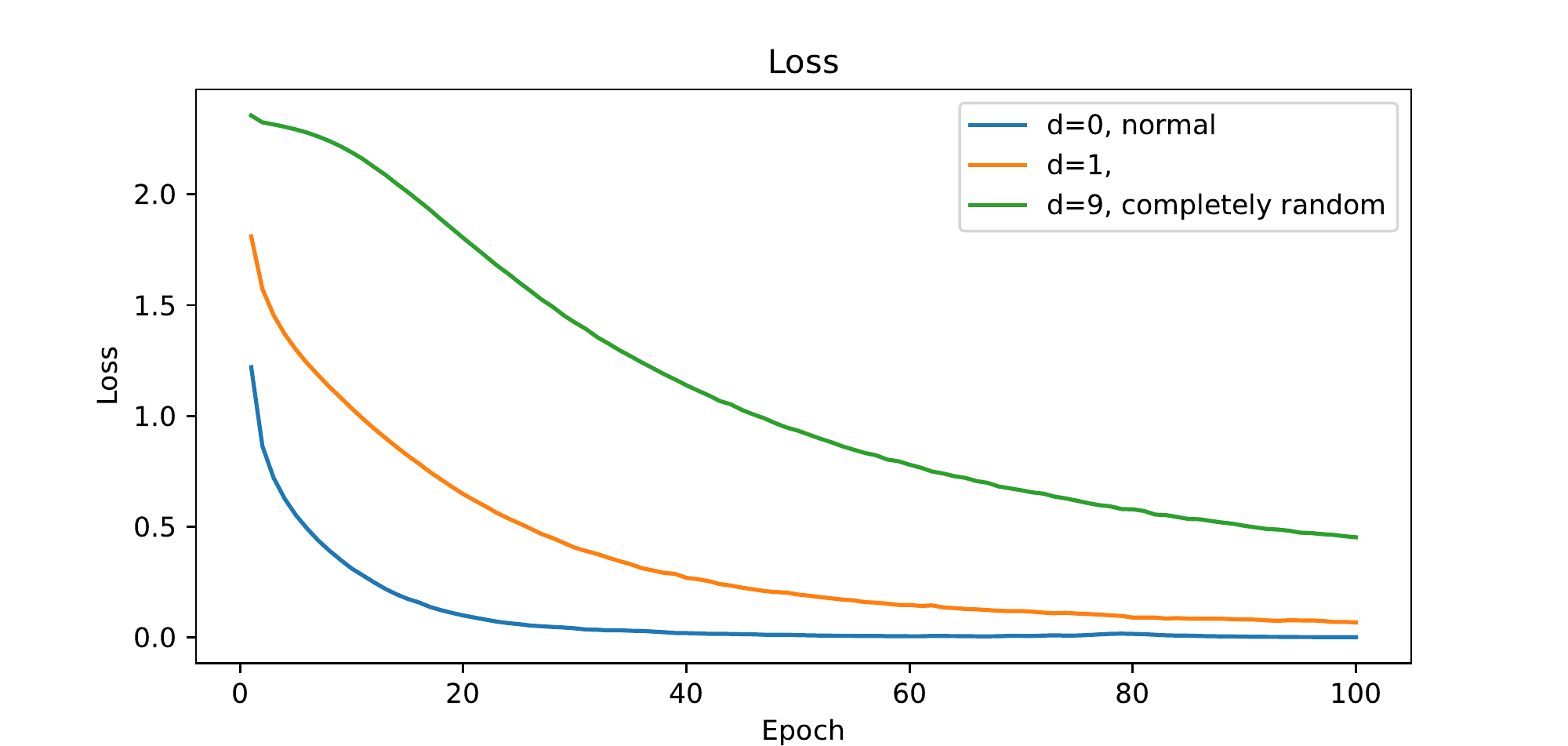}}
\subfigure[]{\label{fig:train}\includegraphics[width=\columnwidth]{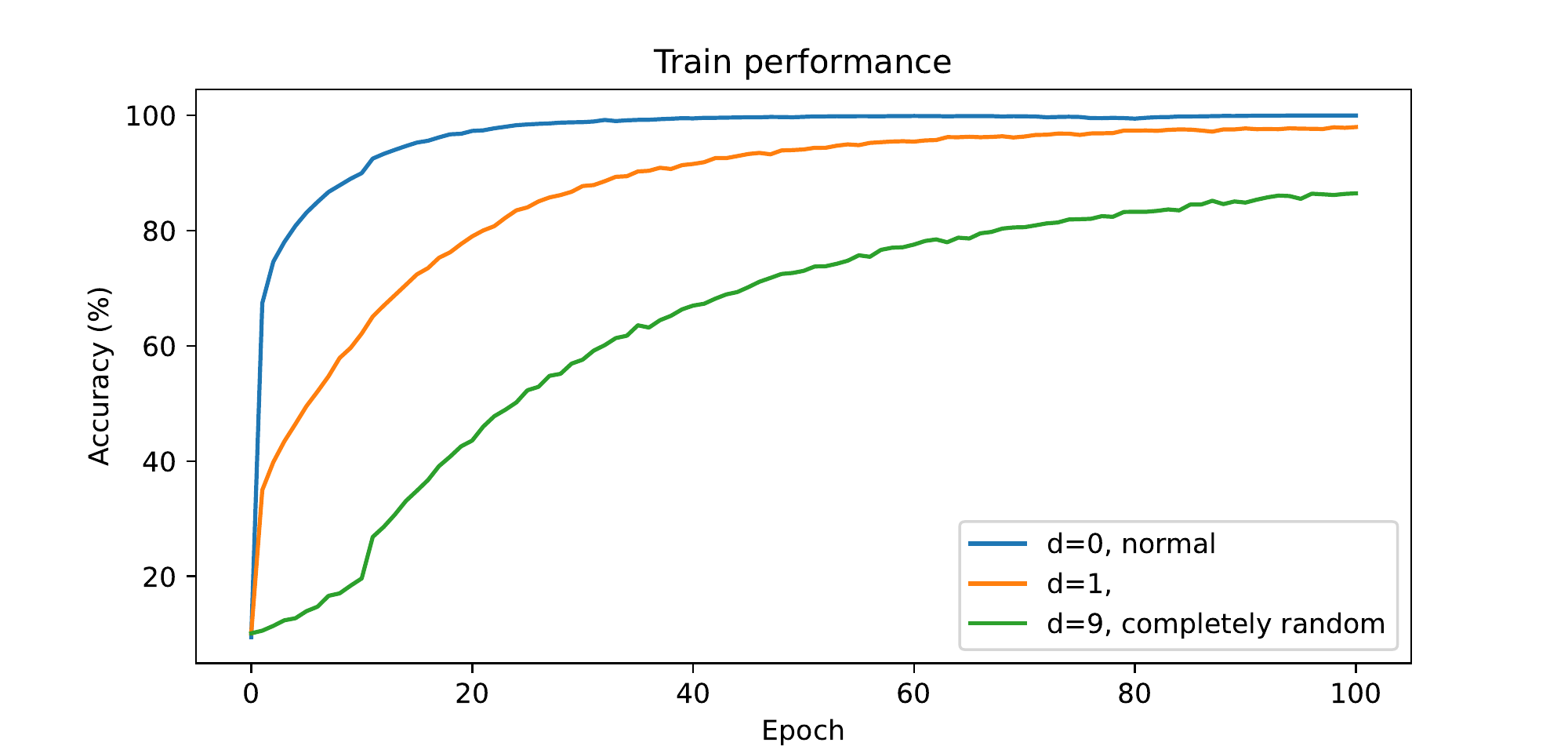}}
\subfigure[]{\label{fig:test}\includegraphics[width=\columnwidth]{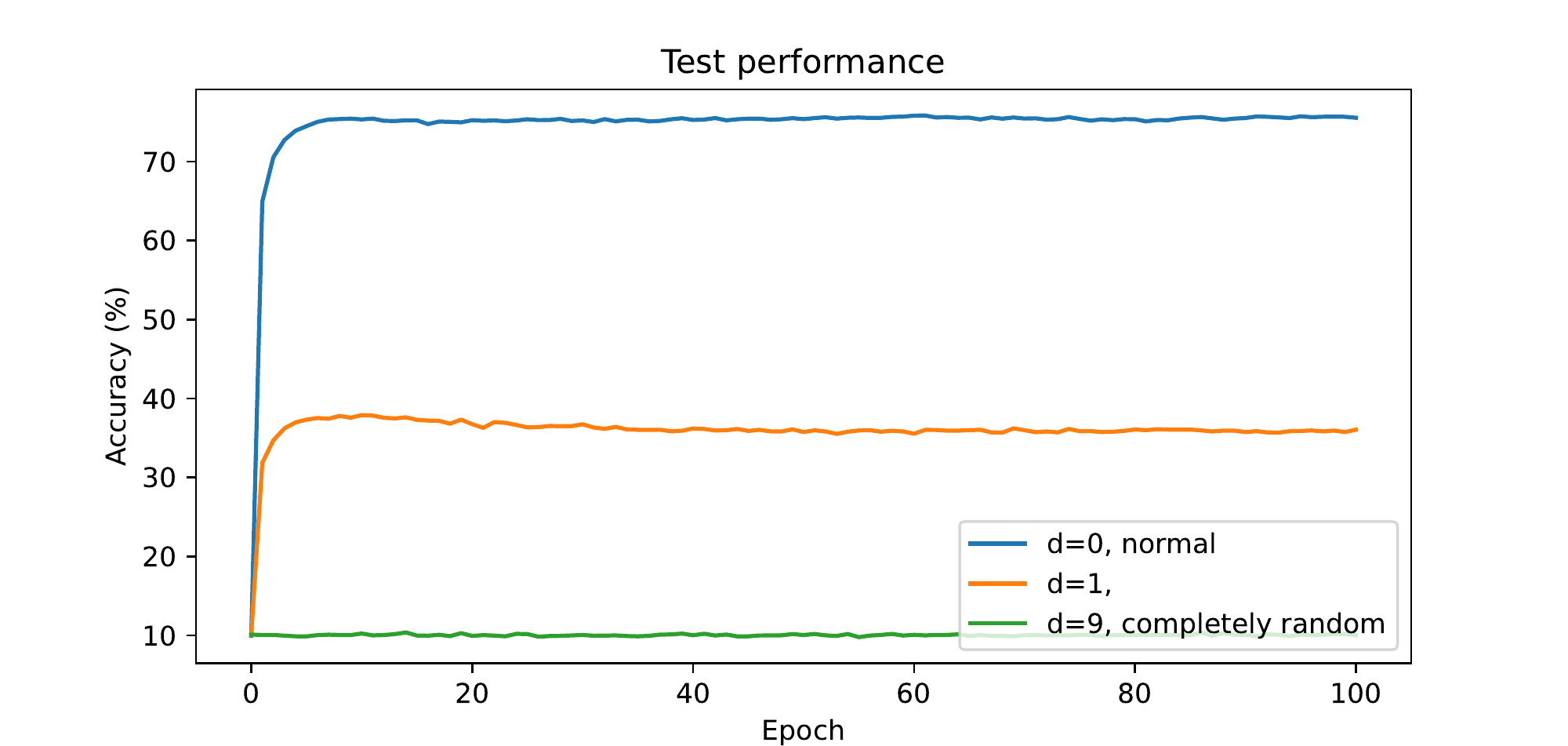}}
\caption{This figure depicts the training statistics of experiments on CIFAR-10 with (partially) random labels: in (a) the loss, in (b) the training accuracy and in (c) the test accuracy.}
\label{fig:stats}
\end{figure}

\subsection{Decrease of Structure in Training Data}\label{sec:deacreas}

Before we analyze the similarity, we want to look at the training statistics, which are depicted in Figure~\ref{fig:stats}.
First of all, in Figure~\ref{fig:loss} be seen that if the data is more random and therefore has less structure, the loss decreases slower. This makes sense, as the overall problem is more difficult, and it has to memorize the labels and cannot take advantage of the shared structure in the data.
Nevertheless, if the models are trained long enough, the training accuracy in the experiments on the randomized data is almost as high as when trained on the randomized data (see  Figure~\ref{fig:train}). It is really interesting that the same neural network on the one side can generalize if data is structured, but on the other side, memorize if the data does not allow to learn structure. This even holds when the networks are largely overparameterized (\citet{DBLP:journals/corr/abs-1801-00173}).
As a sanity check, we depict the test accuracy in Figure~\ref{fig:test}, which of course, only learns something when there is structure. The test accuracy on CIFAR-10 is not matching the state-of-the-art. However, the focus of the paper is not to beat it but to understand representation change in transfer learning.

\begin{figure}[ht]\label{}
\centering     
\subfigure[]{\label{fig:random}\includegraphics[width=0.49\columnwidth]{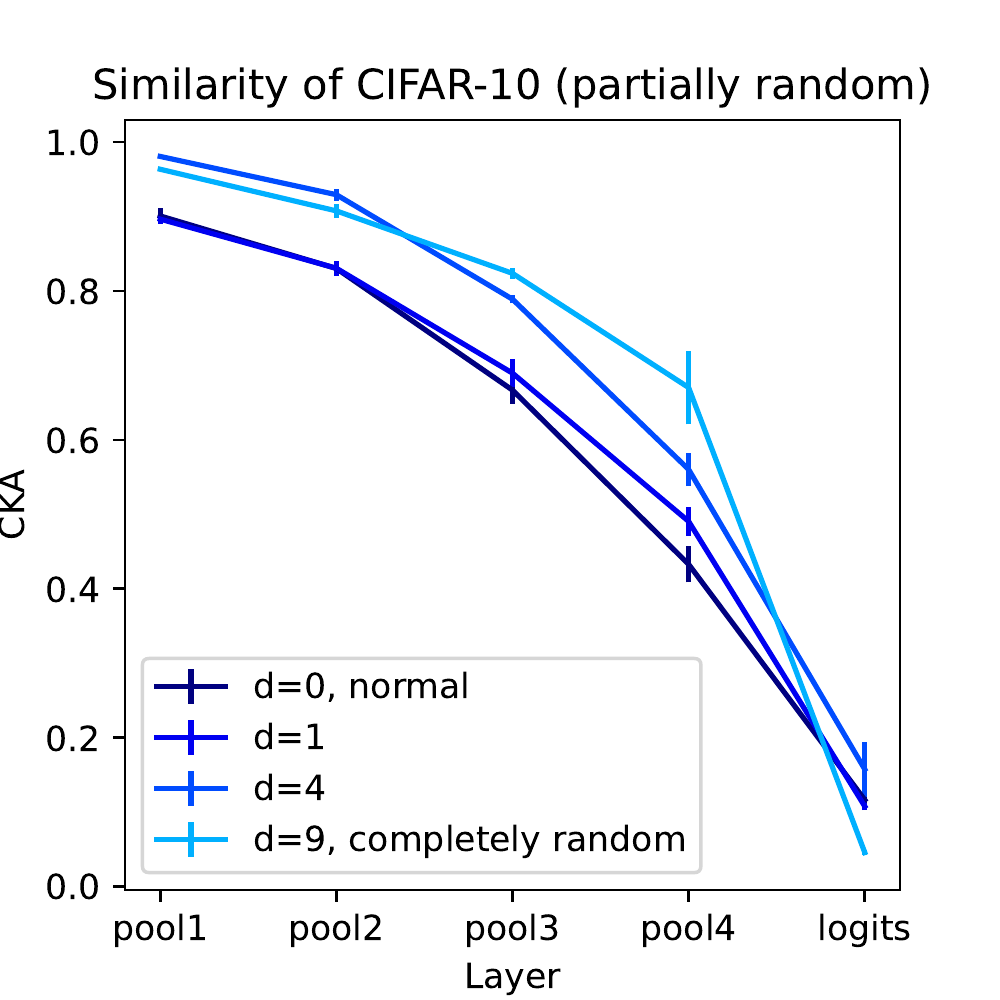}}
\subfigure[]{\label{fig:size}\includegraphics[width=0.49\columnwidth]{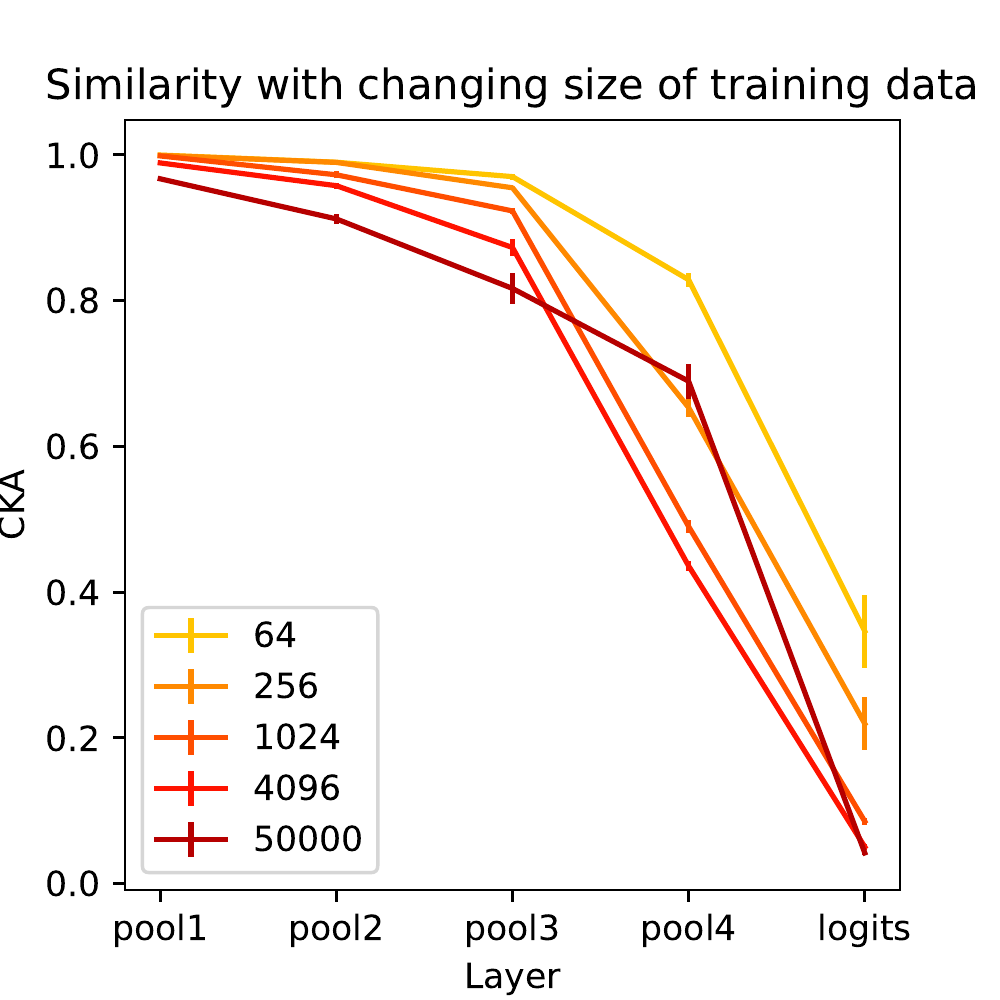}}
\caption{Comparision of the similarity of the experiment on CIFAR-10 with (partially) random labels (\ref{fig:random}). On Figure~\ref{fig:size} the change of representation can be analyzed with respect to the numbers of sample in training size for random labels.}
\end{figure}
Looking at the results of Figure~\ref{fig:random}, it can be observed that if the structure in the data decreases, there is less similarity change overall and especially in the early layers. From that, we conclude that memorization is mainly happening in the last layer. However, the clearer the structure is in the training set, the more the earlier layers are used.

We are also interested in how the representations change with the size of the data when training on completely random labels. In Figure~\ref{fig:size} we see that if there are only a few samples, memorization does not need anything else than the last layer. The more data we have, the more are also earlier layers used.

Overall, we can summarize so far that a larger change in early layers coheres with the learned structure.

\section{Fine-tuned Networks}

Instead of using randomly initialized weights, transfer learning takes advantage of pre-trained weights. In this section, we analyze the representation change of models during fine-tuning.
For all the following comparisons, we pre-trained all models on the training set of CIFAR-10 (5 different seeds).
We fine-tuned several versions of the CIFAR-10 test data and the SVHN dataset.
next to (partially) random versions (see Section~\ref{sec:pre-trained}), we also shifted the labels, such the new labels $Y_s$ are
\begin{equation}
y_S = (y + s) \mod D
\end{equation}
where $s \in {1, ..., D-1}$. This represents a similar dataset that lies in the same domain, but the novel labels have to be relearned.
For all fine-tuned tasks, we used $5000$ samples for training and $5000$ for testing\footnote{also in SVHN, although there is, in theory, more data available, we want to be consistent}.

\begin{figure}[ht]\label{fig:trans_random}
\centering     
\subfigure[]{\label{fig:trans_random}\includegraphics[width=0.49\columnwidth]{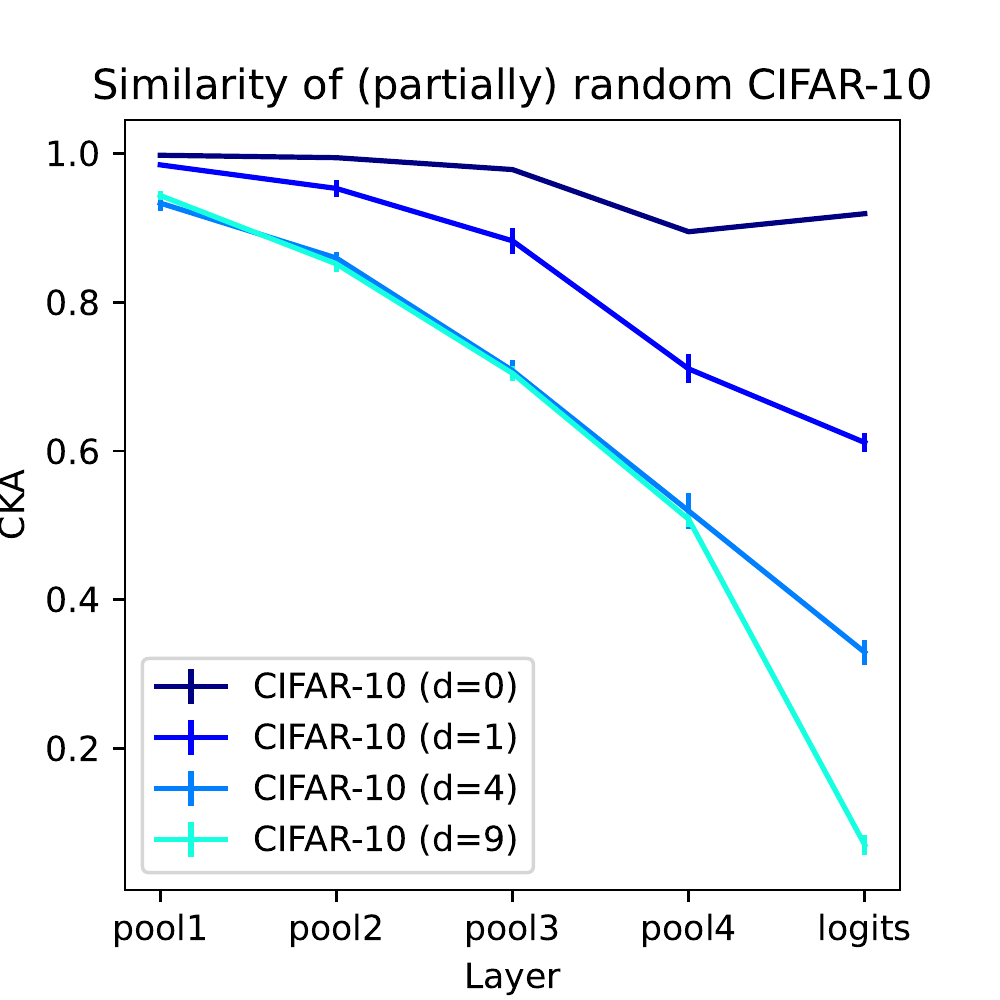}}
\subfigure[]{\label{fig:svhn}\includegraphics[width=0.49\columnwidth]{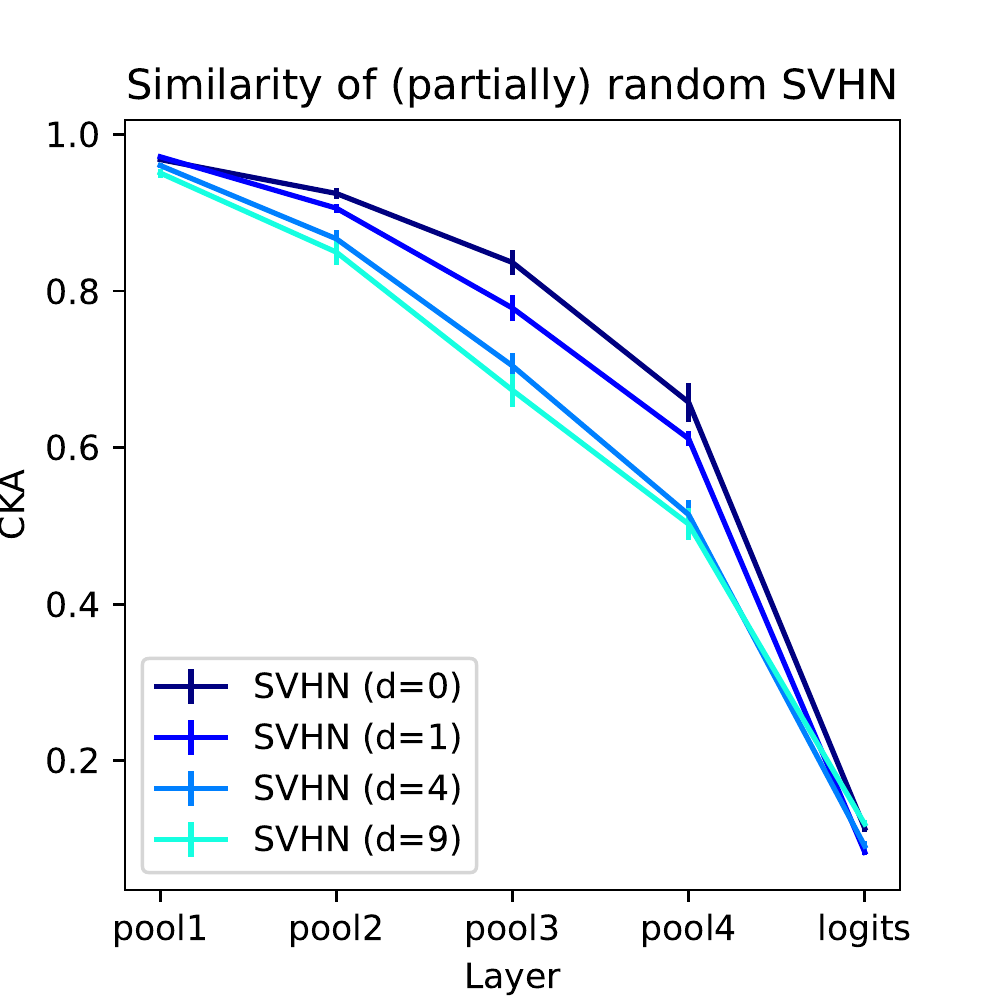}}
\subfigure[]{\label{fig:shifted}\includegraphics[width=0.49\columnwidth]{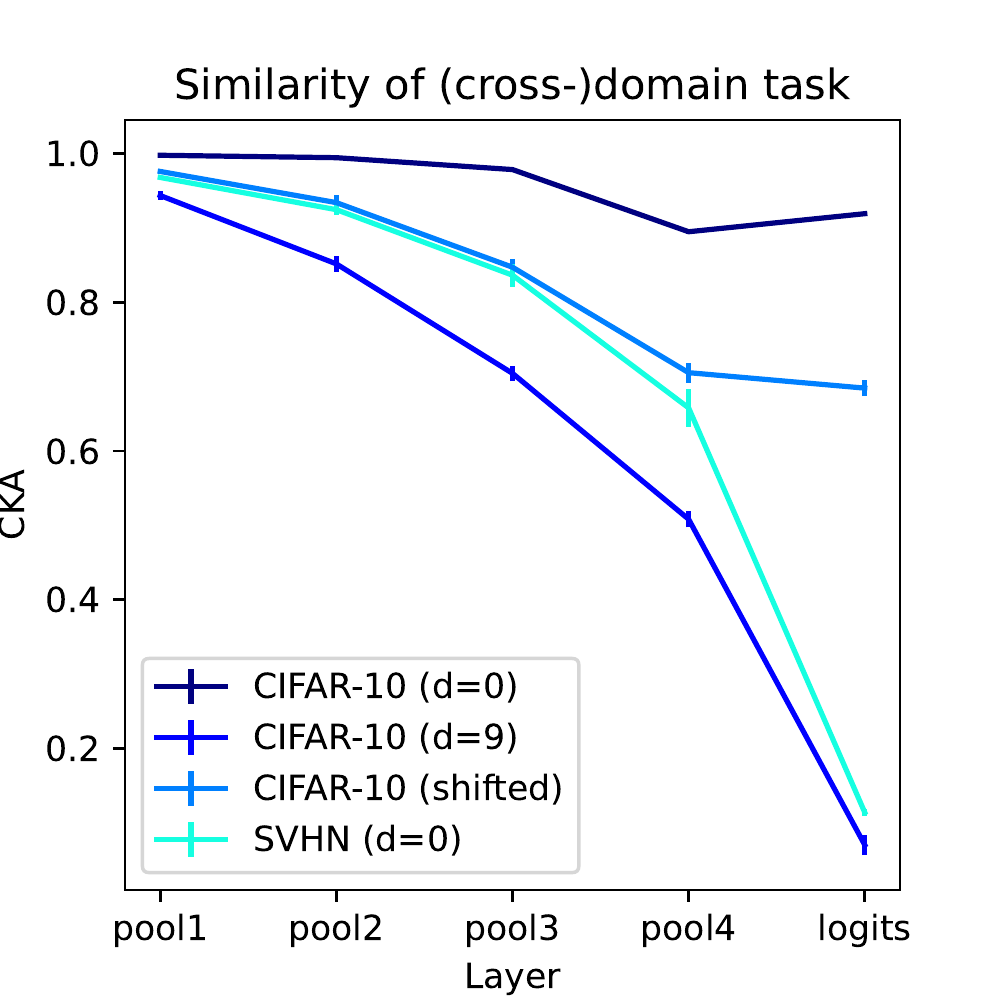}}
\caption{Representation change of fine-tuned networks, which were initilized by pre-trained weights.}
\end{figure}

\subsection{(Partially) Random Labels}

Looking at the results of fine-tuning the (partially) random labels on CIFAR-10 (Figure~\ref{fig:trans_random}), the opposite of the results in standard deep learning happens (see Section~\ref{sec:deacreas}). The more random the data is, the more the representation changes. It makes sense that for non-random labels, hardly anything changes in early layers as the data from the problem comes from the same distribution as in pre-training. However, the learned structure is \textit{unlearned} when there is no structure in the dataset because the labels are random. Interestingly the representation from the complete random experiment is higher than when it was trained on random weights.
We repeated the experiment also on SVHN and a randomized version of it (Figure~\ref{fig:svhn}) to shift the data distribution. But also here, the same is observed, and the more randomness, the more change in representation. 

Therefore, we propose that if a model has learned a specific structure of the data, this is only used when the structure is actually needed.

\subsection{Domain vs. Cross-domain}

In Figure~\ref{fig:shifted}, the results of fine-tuning on a domain (shifted) and a cross-domain (SVHN) tasks are depicted.
The representation change in early layers is larger than when fine-tuned on CIFAR-10 but significantly less than when fine-tuned on random labels. This indicates that the models make use of pre-trained weights. Compared to domain adaption, a cross-domain adaptation relies more on change in the penultimate layer, which has already been mentioned by \citet{DBLP:conf/iclr/OhYKY21}.

\subsection{Pre-trained vs Pre-initilized}

\begin{figure}[ht]\label{fig:preinit3}
\centering     
\subfigure[]{\label{fig:a}\includegraphics[width=0.49\columnwidth]{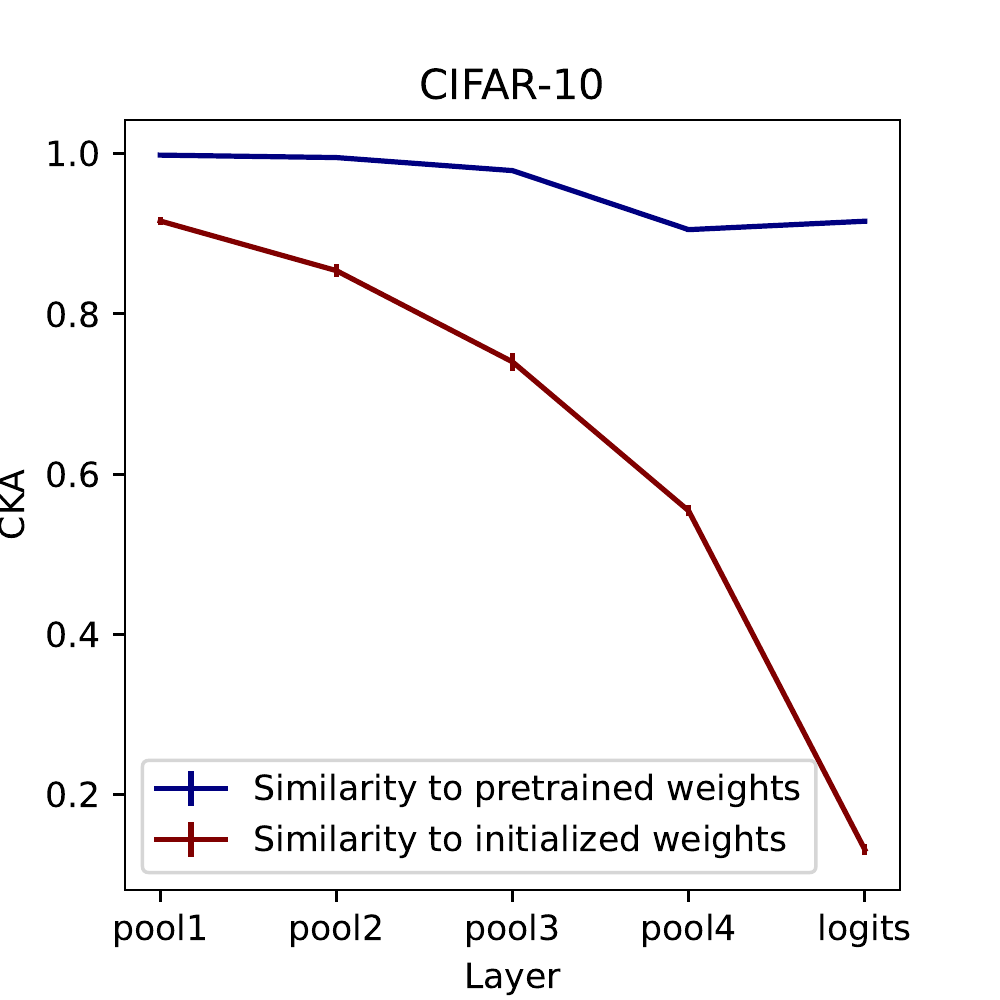}}
\subfigure[]{\label{fig:b}\includegraphics[width=0.49\columnwidth]{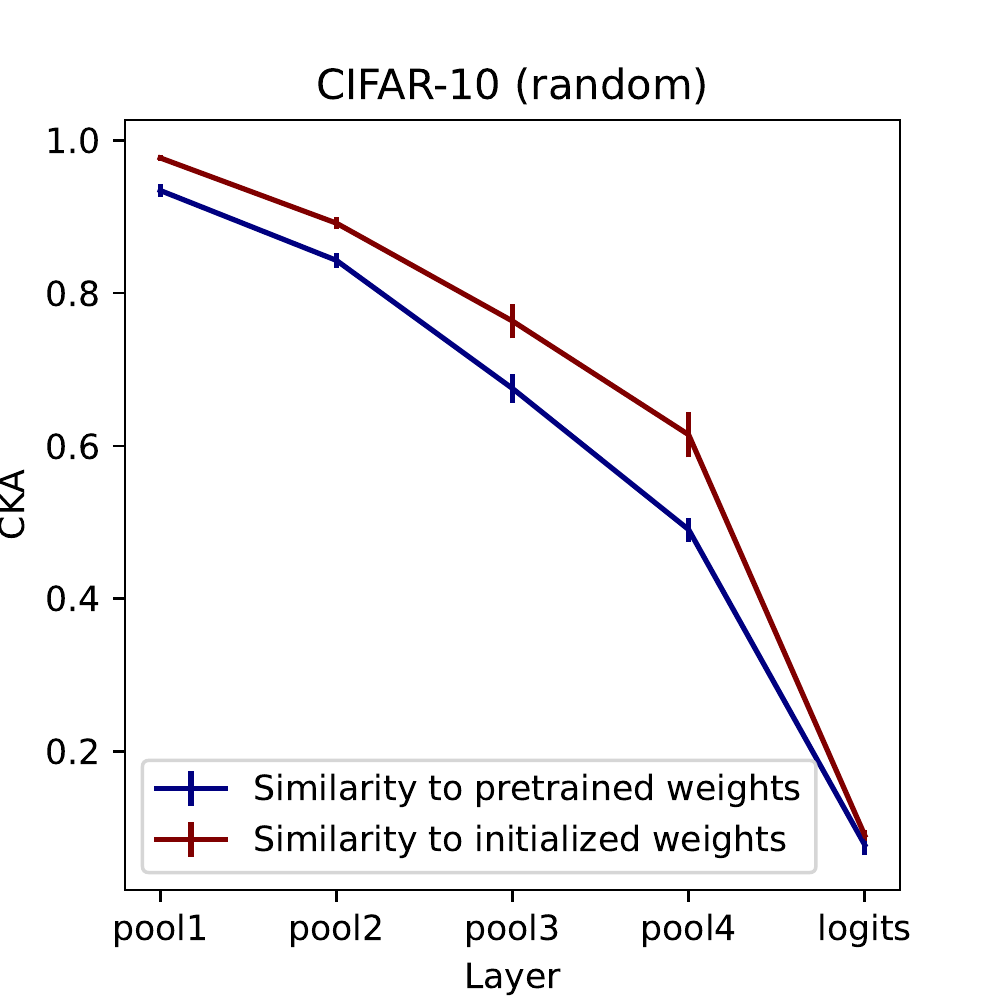}}
\subfigure[]{\label{fig:c}\includegraphics[width=0.49\columnwidth]{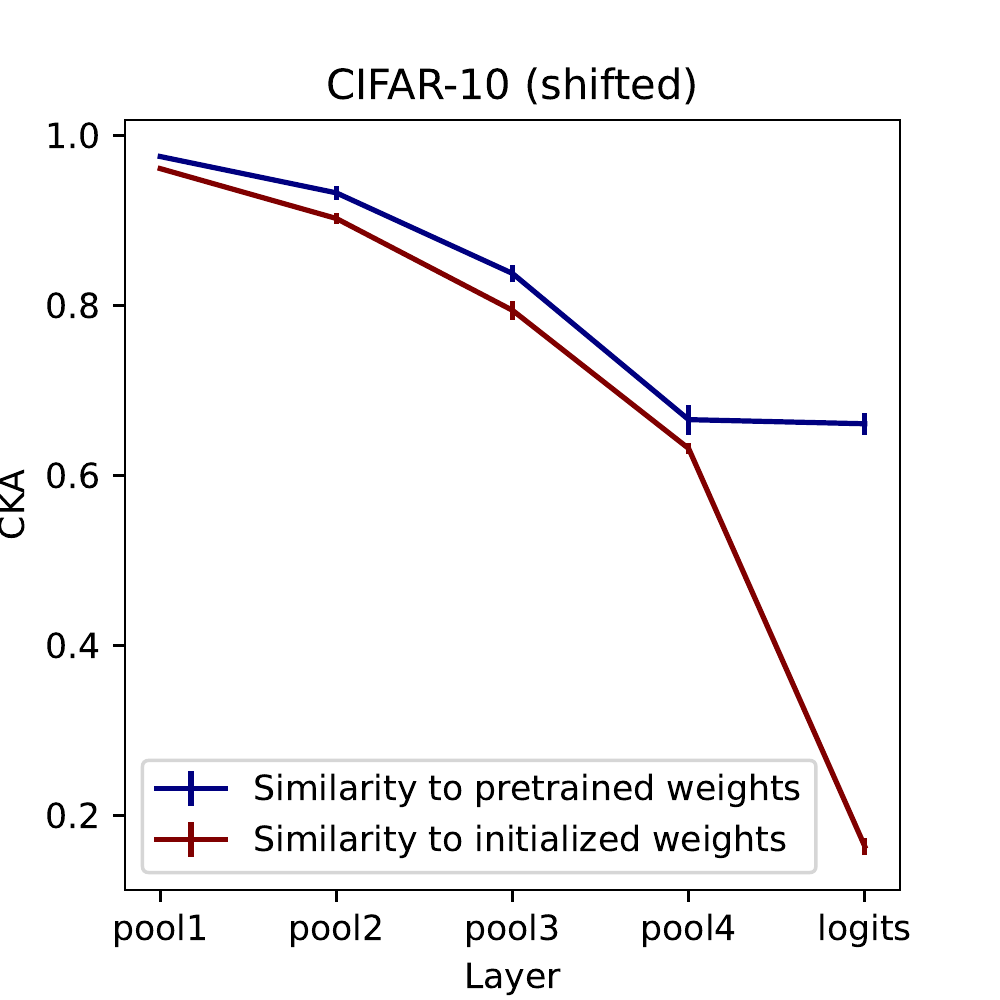}}
\subfigure[]{\label{fig:d}\includegraphics[width=0.49\columnwidth]{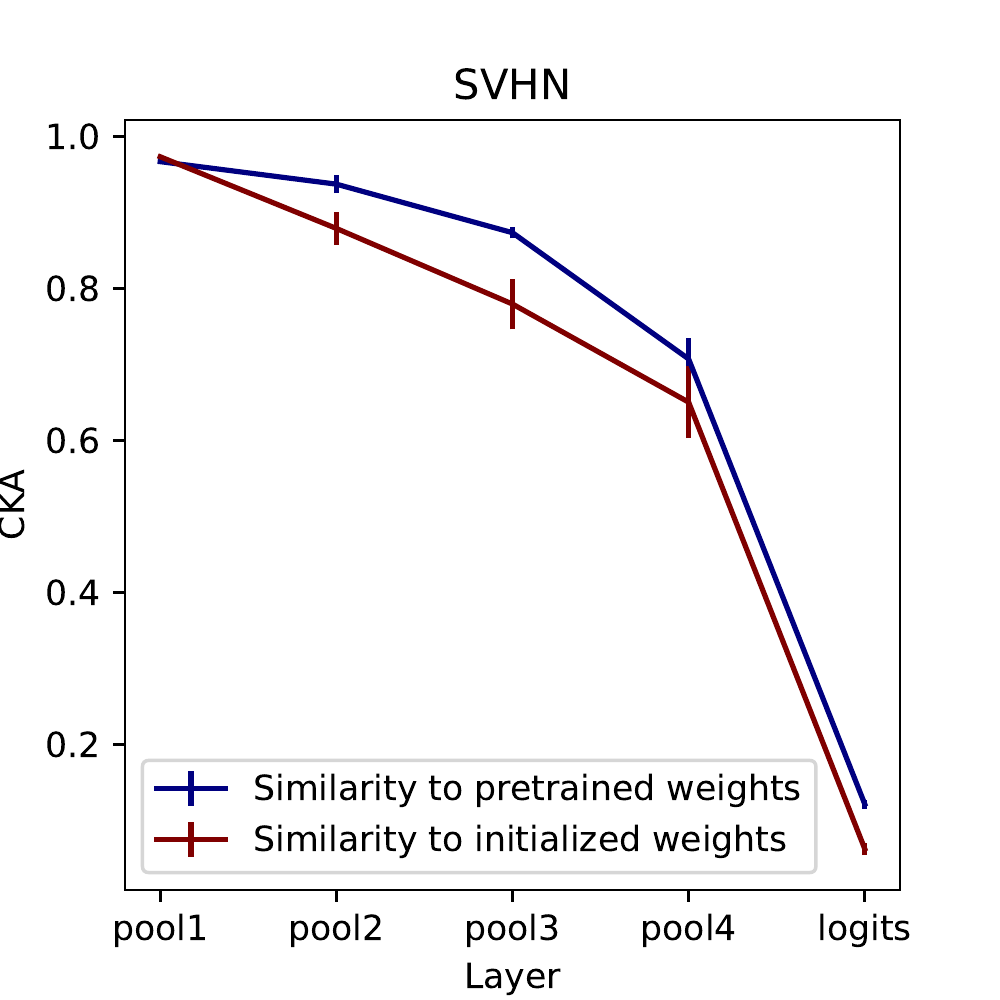}}
\caption{Comparision of the representation change with respect to the representations of initialized and pre-trained weights.}
\end{figure}

Our models are trained in two phases. In the first phase, it transforms the random weights into pre-trained ones, and in the second, it turns the pre-trained weights into fine-tuned ones. In these experiments, we compare the representation change of the fine-tuned weights to both the random initialization and the representation obtained by the pre-trained weights.

In Figure~\ref{fig:a}, we can observe that for CIFAR-10, the fine-tuned representations are closer to the pre-trained than to the pre-initialized one. This makes sense, as the task during pre-training was on data coming from the same distribution, so very similar. Interestingly, when having complete random labels, this is different (see Figure~\ref{fig:b}). The representations obtained after applying the final model to the data are more similar to the complete random initialization than to the pre-trained weights on which the fine-tune task started. This confirms our claim that if the data is not meaningful and structured directly, it \textit{unlearns} learned structure, as this is not helpful in memorization.
However, when the model has learned the structure, it helps to fine-tune and is strongly used when applied to a task that can take advantage of it. In Figure~\ref{fig:c} and Figure~\ref{fig:d}, we see that the fine-tune task exploits the pre-trained weights when applied not on almost the same, but on a domain, respectively, a cross-domain task. This is the case because the final representation is more similar to the pre-trained one than to the initial one, although the differences between the two similarities are not as far away as when the task is almost the same. From the results on random labels (Figure~\ref{fig:b}), we know that this is not automatically the case. The tasks have to share some underlying structure, which goes beyond the similar input in the input layers. Even if the input is the same, random weights destructure the dataset such that pre-trained encoded information is not useful.

\section{Conclusion}

In this paper, we did several experiments to understand how transfer learning takes advantage of pre-trained weights. We used CKA to analyze the representation and reveal the representation change during pre-training and fine-tuning. We found out that memorization happens mainly in the last layer, if possible. Early layers do not change. However, if memorization is learned on a pre-trained model, it \textit{unlearns} the structure and changes the early representation. Domain and cross-domain tasks exploit pre-trained representation and take advantage of it. When solving cross-domain tasks, the earlier layers have to change more than when solving domain tasks to adapt better to the new input data.

\bibliography{references}
\bibliographystyle{icml2022}

%
%
%
\end{document}